\documentclass[journal]{article}

\usepackage{microtype}
\usepackage{graphicx}
\usepackage[margin=1in]{geometry}
\usepackage{caption}
\usepackage{subcaption}
\usepackage{booktabs} 
\usepackage{xurl}
\usepackage{algorithm}
\usepackage{algorithmic}
\usepackage{cite}
\usepackage{hyperref}

\usepackage[normalem]{ulem}
\usepackage{url,multirow,comment}

\usepackage{amsmath,amssymb,amsthm}
\usepackage{graphicx,color}
\usepackage[figuresright]{rotating}
\usepackage[utf8]{inputenc} 
\usepackage[T1]{fontenc}    
\usepackage{url}            
\usepackage{booktabs}       
\usepackage{enumitem}
\usepackage{amsfonts}       
\usepackage{amssymb,mathtools}
\usepackage{amsfonts}
\usepackage{nicefrac}       
\usepackage{microtype}      
\usepackage{amsmath,amssymb,amsthm}
\usepackage{pbox}
\usepackage{enumitem}
\usepackage{url}
\usepackage{bm}

\newtheorem{thm}{Theorem}
\newtheorem{defn}{Definition}

\newtheorem{assm}{Assumption}

\newtheorem{propty}{Property}
\newtheorem{rem}{Remark}

\makeatletter
\def\thanks#1{\protected@xdef\@thanks{\@thanks
        \protect\footnotetext{#1}}}
\makeatother

\begin{document}

\title{Robust Counterfactual Explanations for Tree-Based Ensembles}
\date{}

\author{Sanghamitra Dutta, Jason Long, Saumitra Mishra, Cecilia Tilli, Daniele Magazzeni\\
\normalsize JP Morgan AI Research
\thanks{Accepted at ICML 2022.}
\thanks{The authors are affiliated with JP Morgan AI Research. Author contacts: Sanghamitra Dutta (sanghamitra2612@gmail.com), Jason Long (jason.x.long@jpmorgan.com), Saumitra Mishra (saumitra.mishra@jpmorgan.com), Cecilia Tilli (cecilia.tilli@jpmorgan.com), Daniele Magazzeni (daniele.magazzeni@jpmorgan.com).}}
\maketitle

\begin{abstract}
Counterfactual explanations inform ways to achieve a desired outcome from a machine learning model. However, such explanations are not robust to certain real-world changes in the underlying model (e.g., retraining the model, changing hyperparameters, etc.), questioning their reliability in several applications, e.g., credit lending. In this work, we propose a novel strategy -- that we call \emph{RobX} -- to generate robust counterfactuals for tree-based ensembles, e.g., XGBoost. Tree-based ensembles pose additional challenges in robust counterfactual generation, e.g., they have a non-smooth and non-differentiable objective function, and they can change a lot in the parameter space under retraining on very similar data.  We first introduce a novel metric -- that we call \emph{Counterfactual Stability} -- that attempts to quantify how robust a counterfactual is going to be to model changes under retraining, and comes with desirable theoretical properties. Our proposed strategy \emph{RobX} works with any counterfactual generation method (base method) and searches for robust counterfactuals by iteratively refining the counterfactual generated by the base method using our metric \emph{Counterfactual Stability}. We compare the performance of \emph{RobX} with popular counterfactual generation methods (for tree-based ensembles) across benchmark datasets. The results demonstrate that our strategy generates counterfactuals that are significantly more robust (nearly $100\%$ validity after actual model changes) and also realistic (in terms of local outlier factor) over existing state-of-the-art methods.
\end{abstract}

\parindent 0pt
\topsep 4pt plus 1pt minus 2pt
\partopsep 1pt plus 0.5pt minus 0.5pt
\itemsep 2pt plus 1pt minus 1pt
\parsep 2pt plus 1pt minus 0.5pt
\parskip 6pt

\section{Introduction}
\label{sec:introduction}
Counterfactual explanations have generated immense interest in several high-stakes applications, e.g., lending, credit decision, hiring, etc~\cite{verma2020counterfactual, Karimi_arXiv_2020,wachter2017counterfactual}. Broadly speaking, the goal of counterfactual explanations is to guide an applicant on how they can change the outcome of a model by providing suggestions for improvement. Given a specific input value (e.g., a data point that is declined by a model), counterfactual explanations attempt to find another input value for which the model would provide a different outcome (essentially get accepted). Such an input value that changes the model outcome is often referred to as a \emph{counterfactual}.

Several existing works usually focus on finding counterfactuals that are as ``close'' to the original data point as possible with respect to various distance metrics, e.g., $L_1$ cost or $L_2$ cost. This cost is believed to represent the ``effort'' that an applicant might need to make to get accepted by the model. Thus, the ``closest'' counterfactuals essentially represent the counterfactuals attainable with minimum effort.


However, the closest counterfactuals may not always be the most preferred one. For instance, if the model changes even slightly, e.g., due to retraining, the counterfactual may no longer remain valid. In Table~\ref{robustness_demo}, we present a scenario where we retrain an XGBoost model~\cite{chen2015xgboost} with same hyperparameters on the same dataset, leaving out just one data point. We demonstrate that a large fraction of the ``closest'' counterfactuals generated using the state-of-the-art techniques for tree-based models no longer remain valid. This motivates our primary question: 
\begin{center}
\emph{How do we generate counterfactuals for tree-based ensembles that are not only close but also robust to changes in the model?}
\end{center}

\begin{table}
\caption{Validity of Counterfactuals Generated Using State-Of-The-Art Techniques (with $L_1$ cost minimization) for XGBoost Models on German Credit Dataset~\cite{UCI}: Models were retrained after dropping only a single data point. A large fraction of the counterfactuals for the previous model no longer remain valid for the new models obtained after retraining.}
\label{robustness_demo}
\begin{center}
\begin{small}
\begin{sc}
\begin{tabular}{lcccc}
\toprule
Method & FT & FOCUS & FACE & NN  \\
\midrule
Validity   & $72.9\%$  
  & $72.8\%$ 
  & $84.4\%$  
  &  $92.5\%$\\  
\bottomrule
\end{tabular}
\end{sc}
\end{small}
\end{center}
\end{table}

Towards addressing this question, in this work, we make the following contributions:

\begin{itemize}
\item \textbf{Quantification of Counterfactual Stability:} We propose a novel metric that we call -- \emph{Counterfactual Stability} -- that quantifies how robust a counterfactual is going to be to possible changes in the model. In order to arrive at this metric, we identify the desirable theoretical properties of counterfactuals in tree-based ensembles that can make them more stable, i.e., less likely to be invalidated due to possible model changes under retraining. Tree-based ensembles pose additional challenges in robust counterfactual generation because they do not conform to standard assumptions, e.g., they are not smooth and continuous, have a non-differentiable objective function, and can change a lot in the parameter space under retraining on similar data. Our proposed quantification is of the form $R_{\Phi}(x,M)$ where $x \in \mathbb{R}^d$ is an input (not necessarily in the dataset or data manifold), $M(\cdot):\mathbb{R}^d \to [0,1]$ is the original model, and $\Phi$ denotes some hyperparameters for this metric. We find that while counterfactuals on the data manifold have been found to be more robust than simply ``closest'' or ``sparsest'' counterfactuals (see \cite{pawelczyk2020counterfactual}), being on the data manifold may not be sufficient for robustness, thus calling for our metric.

\item \textbf{Conservative Counterfactuals With Theoretical Robustness Guarantee:}
We introduce the concept of \emph{Conservative Counterfactuals} which are essentially counterfactuals (points with desired outcome) lying in the dataset that also have high counterfactual stability $R_{\Phi}(x,M)$. Given an input $x \in \mathbb{R}^d$, a conservative counterfactual is essentially its nearest neighbor in the dataset on the other side of the decision boundary that also passes the counterfactual stability test, i.e., $R_{\Phi}(x,M) \geq \tau$ for some threshold $\tau$. We provide a theoretical guarantee (see Theorem~\ref{thm:guarantee}) that bounds the probability of invalidation of the conservative counterfactual under model changes.

\item \textbf{An Algorithm for Robust Counterfactual Explanations (RobX):} We propose \emph{RobX} that generates robust counterfactuals for tree-based ensembles leveraging our metric of counterfactual stability. Our proposed strategy is a post-processing one, i.e., it can be applied after generating counterfactuals using any of the existing methods for tree-based ensembles (that we also refer to as the base method), e.g., Feature Tweaking (FT)~\cite{tolomei2017interpretable}, FOCUS~\cite{lucic2019focus}, Nearest Neighbor (NN)~\cite{albini2021counterfactual}, FACE~\cite{poyiadzi2020face},  etc. Our strategy iteratively refines the counterfactual generated by the base method and moves it towards the conservative counterfactual, until a ``stable'' counterfactual is found (i.e., one that passes our counterfactual stability test $R_{\Phi}(x,M) \geq \tau$).

\item \textbf{Experimental Demonstration:} Our experiments on real-world datasets, namely, German Credit~\cite{UCI}, and HELOC~\cite{Fico_web_2018}, demonstrate that the counterfactuals generated using RobX significantly improves the robustness of counterfactuals over SOTA techniques (nearly 100\% validity after actual model changes). Furthermore, our counterfactuals also lie in the dense regions of the data manifold, thereby being realistic in terms of Local Outlier Factor (see Definition~\ref{defn:lof}), a metric popularly used to quantify likeness to the data manifold. 
\end{itemize}

\begin{figure}
\centering
\includegraphics[height=4.2cm]{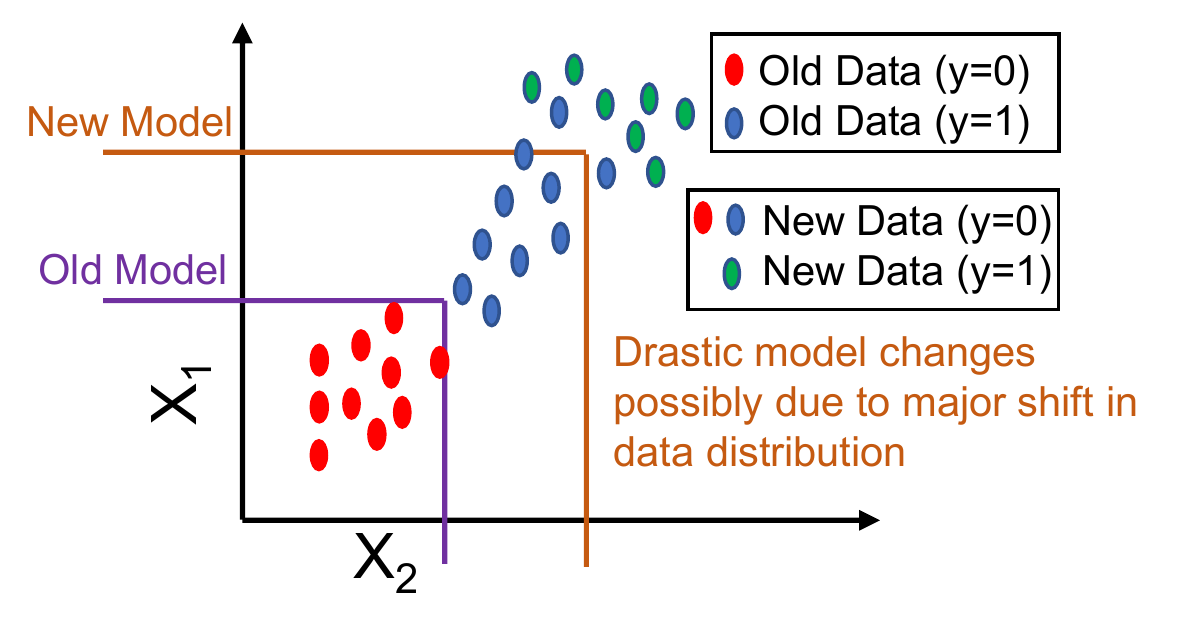}
\includegraphics[height=4.2cm]{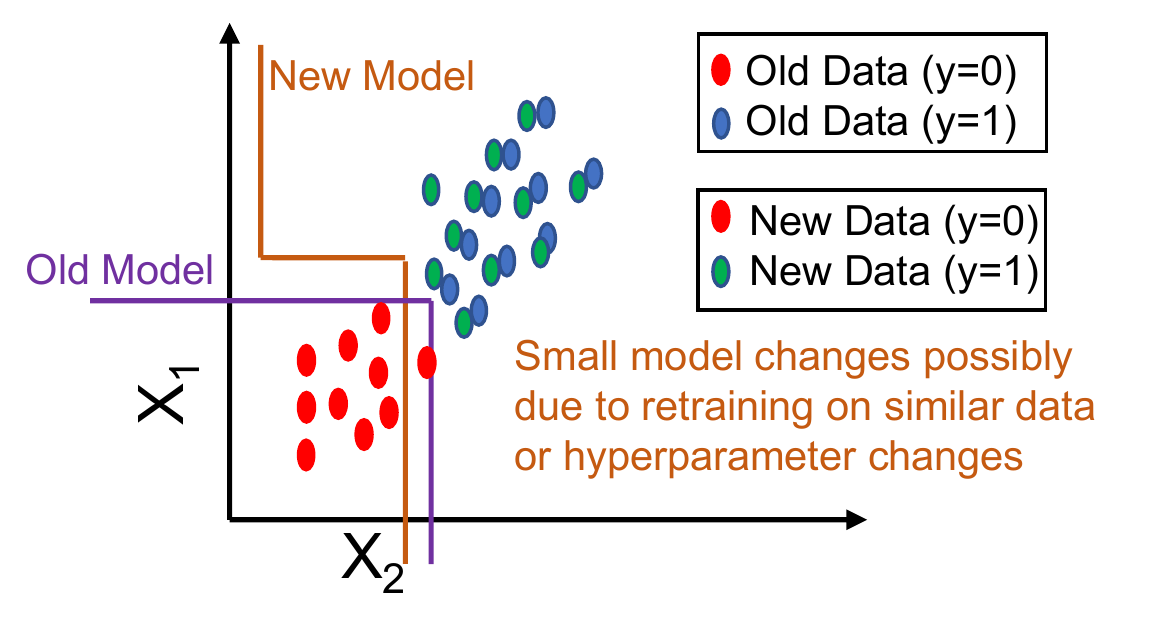}
\caption{Scenarios distinguishing drastic and small model changes: (Left) Drastic model changes due to major distribution shifts; One may not want robustness of counterfactuals here. (Right) Small model changes due to retraining on very similar data or minor hyperparameter changes that occur frequently in practice. Robustness of counterfactuals is highly desirable here.}
\label{fig:robustness_not_needed}
\end{figure}

\begin{rem}[Drastic Model Changes]
One might question that why should one want counterfactuals to necessarily remain valid after changes to the model. Shouldn't they instead vary with the model to reflect the changes to the model? E.g., economic changes might cause drastic changes in lending models (possibly due to major data distribution shifts). In such scenarios, one might in fact prefer counterfactuals for the old and new models to be different. Indeed, we agree that counterfactuals are not required to remain valid for very drastic changes to the model (see Figure~\ref{fig:robustness_not_needed}; also see an impossibility result in Theorem~\ref{thm:impossibility1}). However, this work focuses on small changes to the model, e.g., retraining on some data drawn from the same distribution, or minor changes to the hyperparameters, keeping the underlying data mostly similar. Such small changes to the model are in fact quite common in several applications and occur frequently in practice~\cite{upadhyay2021towards,Hancox-Li_fat_2020,black2021consistent,barocas2020hidden}.
\end{rem}

\textbf{Related Works:} Counterfactual explanations have received significant attention in recent years (see \cite{verma2020counterfactual,Karimi_arXiv_2020,wachter2017counterfactual,multiobjective,konig2021causal,albini2021counterfactual,kanamori2020dace,poyiadzi2020face,lucic2019focus,pawelczyk2020counterfactual,ley2022global,spooner2021counterfactual,sharma2019certifai} as well as the references therein). In \cite{pawelczyk2020counterfactual,kanamori2020dace,poyiadzi2020face}, the authors argue that counterfactuals that lie on the data manifold are likely to be more robust than the closest counterfactuals, but the focus is more on generating counterfactuals that specifically lie on the data manifold (which may not always be sufficient for robustness). Despite researchers arguing that robustness is an important desideratum of local explanation methods~\cite{Hancox-Li_fat_2020}, the problem of generating robust counterfactuals has been less explored, with the notable exceptions of some recent works \cite{upadhyay2021towards,rawal2020can,black2021consistent}. In \cite{upadhyay2021towards,black2021consistent}, the authors propose algorithms that aim to find the \emph{closest} counterfactuals that are also robust (with demonstration on linear models and neural networks). In \cite{rawal2020can}, the focus is on analytical trade-offs between validity and cost. We also refer to \cite{Mishra_arXiv_2021} for a survey on the robustness of both feature-based attributions and counterfactuals. 

In this work, our focus is on generating robust counterfactuals for tree-based ensembles. Tree-based ensembles pose additional challenges in robust counterfactual generation because they do not conform to standard assumptions for linear models and neural networks, e.g., they have a non-smooth, non-differentiable objective function. Furthermore, our performance metrics include both distance ($L_1$ or $L_2$ cost), and likeness to the data manifold (LOF). 

We note that \cite{alvarez2018robustness} proposes an alternate perspective of robustness in explanations called $L$-stability which is built on similar individuals receiving similar explanations. Instead, our focus is on explanations remaining valid after some changes to the model.

\section{Problem Setup}
Let $\mathcal{X} \subseteq \mathbb{R}^d$ denote the input space and let $\mathcal{S}=\{x_i\}_{i=1}^N \in \mathcal{X}$ be a dataset consisting of $N$ independent and identically distributed points generated from a density $q$ over $\mathcal{X}$. 
We also let $M (\cdot):\mathbb{R}^d \to [0,1]$ denote the original machine learning model (a tree-based ensemble, e.g., an XGBoost model) that takes an input value and produces an output probability lying between $0$ and $1$. The final decision is denoted by: $$D(x)=\begin{cases} 1 \text{ if $M(x)>0.5$,}\\
0 \text{ otherwise.}\end{cases}$$

Similarly, we denote a changed model by $M_{new}(\cdot):\mathbb{R}^d \to [0,1]$, and the decision of the changed model by: $$D_{new}(x)=\begin{cases} 1 \text{ if $M_{new}(x)>0.5$,}\\
0 \text{ otherwise.}\end{cases}$$

In this work, we are mainly interested in tree-based ensembles~\cite{chen2015xgboost}. A tree-based ensemble model is defined as follows:
$M(x)=\sum_{t=1}^T m^{(t)}(x)$ where each $m^{(t)}(x)$ is an independent tree with $L$ leaves, having weights $\{w_1,\ldots,w_L\} \in \mathbb{R}$. A tree $m^{(t)}(x)$ maps a data point $x \in \mathbb{R}^d$ to one of the leaf indices (based on the tree structure), and produces an output $w_l \in \{w_1,\ldots,w_L\}$. One may use a \texttt{sigmoid} function~\cite{chen2015xgboost} for the final output to lie in $[0,1]$.

\subsection{Background on Counterfactuals}

Here, we provide a brief background on counterfactuals. 

\begin{defn}[Closest Counterfactual $\mathcal{C}_{p}(x,M)$]
Given $x\in \mathbb{R}^d$ such that $M(x)\leq 0.5$, its closest counterfactual (in terms of $L_p$-norm) with respect to the model $M(\cdot)$ is defined as a point $x'\in \mathbb{R}^d$ that minimizes the $l_p$ norm $||x-x'||_p$ such that $M(x')>0.5$. 
\begin{equation}
\mathcal{C}_{p}(x,M)=\arg \min_{x'\in \mathbb{R}^d} ||x-x'||_p 
\text{ such that } M(x')>0.5. \nonumber
\end{equation}  
\end{defn}

For tree-based ensembles, some existing approaches to find the closest counterfactuals include~\cite{tolomei2017interpretable,lucic2019focus}. When $p=1$, these counterfactuals are also referred to as ``sparse'' counterfactuals in existing literature~\cite{pawelczyk2020counterfactual} because they attempt to find counterfactuals that can be attained by changing as few features as possible (enforcing a sparsity constraint). 

Closest counterfactuals have often been criticized in existing literature~\cite{poyiadzi2020face,pawelczyk2020counterfactual,kanamori2020dace} as being too far from the data manifold, and thus being too unrealistic, and anomalous. This has led to several approaches for generating ``data-support'' counterfactuals that are lie on the data manifold, e.g.,  \cite{kanamori2020dace,albini2021counterfactual,poyiadzi2020face}. Here, we choose one such definition of data-support counterfactual which is essentially the nearest neighbor with respect to the dataset $\mathcal{S}$, that also gets accepted by the model~\cite{albini2021counterfactual}.

\begin{defn}[Closest Data-Support Counterfactual $\mathcal{C}_{p,\mathcal{S}}(x,M)$]
\label{defn:data-support-CF}
Given $x\in \mathbb{R}^d$ such that $M(x)\leq 0.5$, its closest data-support counterfactual $\mathcal{C}_{p,\mathcal{S}}(x,M)$ with respect to the model $M(\cdot)$ and dataset $\mathcal{S}$ is defined as a point $x'\in \mathcal{S}$ that minimizes the $l_p$ norm $||x-x'||_p$ such that $M(x')>0.5$.
\begin{equation}
\mathcal{C}_{p,\mathcal{S}}(x,M)=\arg \min_{x'\in \mathcal{S}} ||x-x'||_p
\text{ such that }{M(x')>0.5}. \nonumber
\end{equation}  
\end{defn}

\begin{rem}[Metrics to Quantify Likeness to Data Manifold] In practice, instead of finding counterfactuals that lie exactly on the dataset, one may use alternate metrics that quantify how alike or anomalous is a point with respect to the dataset. One popular metric to quantify anomality that is also used in existing literature~\cite{pawelczyk2020counterfactual,kanamori2020dace} on counterfactual explanations is Local Outlier Factor (see Definition~\ref{defn:lof}; also see \cite{breunig2000lof}).
\end{rem}

\begin{defn}[Local Outlier Factor (LOF)]
For $x \in \mathcal{S}$, let $N_k(x)$ be its $k$-nearest neighbors (k-NN) in $\mathcal{S}$. The $k$-reachability distance $rd_k$ of $x$ with respect to $x'$
is defined by $rd_k(x, x')= \max\{\Delta(x, x'), d_k(x')\}$, where $d_k(x')$
is the distance $\Delta$ between $x'$ and its the $k$-th nearest instance
on $\mathcal{S}$. The $k$-local reachability density of $x$ is defined by
$lrd_k(x) = |N_k(x)| (
\sum_{x' \in N_k(x)} rd_k(x, x'))^{-1}.$ Then, the
k-LOF of $x$ on $\mathcal{S}$ is defined as follows:
$$q_k(x | \mathcal{S}) = \frac{1}{|N_k(x)|}
\sum_{x' \in N_k(x)}
\frac{lrd_k(x')}{lrd_k(x)}
.$$ Here, $\Delta(x, x')$ is the distance between two $d$-dimensional feature vectors.
\label{defn:lof}
\end{defn}

In this work, we use an existing implementation of computing LOF from \texttt{scikit}~\cite{scikit-lof} that predicts $-1$ if the point is anomalous, and $+1$ for inliers. So, in this work, a high average LOF essentially suggests the points lie on the data manifold, and are more realistic, i.e., \emph{higher is better}.

Next, we introduce our goals.
\subsection{Goals}

Given a data point $x \in \mathcal{X}$ such that $M(x)\leq 0.5$, our goal is to find a counterfactual $x'$ with $M(x')>0.5$ that meets our requirements:
\begin{itemize}[leftmargin=*, itemsep=0pt, topsep=0pt]
\item Close in terms of $L_p$ cost: The point $x'$ is close to $x$, i.e., $||x-x'||_p$ is as low as possible.
\item Robust: The point $x'$ remains valid after changes to the model, i.e., $M_{new}(x')>0.5$.
\item Realistic: The point $x'$ is as similar to the data manifold as possible, e.g., has a high LOF (higher is better).
\end{itemize}

\begin{rem}[Bookkeeping Past Counterfactuals] One possible solution for ensuring the robustness of counterfactuals under model changes could be to keep a record of past counterfactuals. Then, even if there are small changes to the model that can render those counterfactuals invalid, one might still want to accept them because they have been recommended in the past: Ouput $D(x) \text{ if x is a past counterfactual}$ or $D_{new}(x)$ otherwise. However, this approach would require significant storage overhead. Furthermore, there would also be fairness concerns if two data points that are extremely close to each other are receiving the same decision, e.g., one is being accepted because it was a past counterfactual even though the new model rejects it, while the other point is being rejected. 
\end{rem}

\section{Main Results}
\label{sec:main}
In this section, we first identify the desirable properties of counterfactuals in tree-based ensembles that make them more stable, i.e., less likely to be invalidated by small changes to the model. These properties then leads us to propose a novel metric -- that we call \emph{Counterfactual Stability} -- that quantifies the robustness of a counterfactual with respect to possible changes to a model. This metric enables us to arrive at an algorithm for generating robust counterfactuals that can be applied over any base method.

\subsection{Desirable properties of counterfactuals in tree-based ensembles that make them more stable}

In this work, we are interested in finding counterfactuals that are robust to small changes to the model (recall Figure~\ref{fig:robustness_not_needed}), e.g., retraining on some data from the same distribution, or minor changes to the hyperparameters. We note that if the model changes drastically, it might not make sense to expect that counterfactuals will remain valid, as demonstrated in the following impossibility result.  
\begin{thm}[Impossibility Under Drastic Model Changes]
Given a tree-based ensemble model $M(\cdot):\mathbb{R}^d \to [0,1]$, there always exists another tree-based ensemble model $M_{new}(\cdot):\mathbb{R}^d \to [0,1]$ such that all counterfactuals to $M$ with respect to a dataset $\mathcal{S}$ no longer remains valid.
\label{thm:impossibility1}
\end{thm}

Thus, we first need to make some \emph{reasonable} assumptions on how the model changes during retraining, or rather, what kind of model changes are we most interested in. 

In this work, we instead arrive at the following desirable properties of counterfactuals for tree-based ensembles that can make them more stable, i.e., less likely to be invalidated. Our first property is based on the fact that the output of a model $M(x) \in [0,1]$ is expected to be higher if the model has more confidence in that prediction. 

\begin{propty}
For any $x \in \mathbb{R}^d$, a higher value of $M(x)$ makes it less likely to be invalidated due to model changes.
\label{propty:high_confidence}
\end{propty}

\begin{figure}
\centering
\begin{subfigure}[b]{0.48\textwidth}
\centering
{\centering \includegraphics[height=4.1cm]{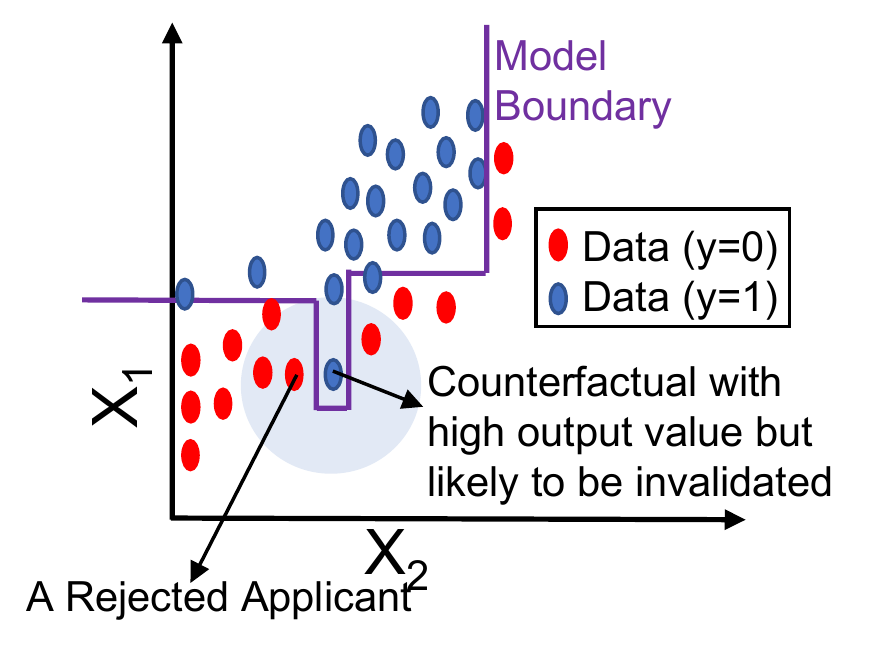}}
\caption{Counterfactual is close to the boundary. \label{fig:property2}}
\end{subfigure}
\begin{subfigure}[b]{0.48\textwidth}
\centering
{\centering \includegraphics[height=4.1cm]{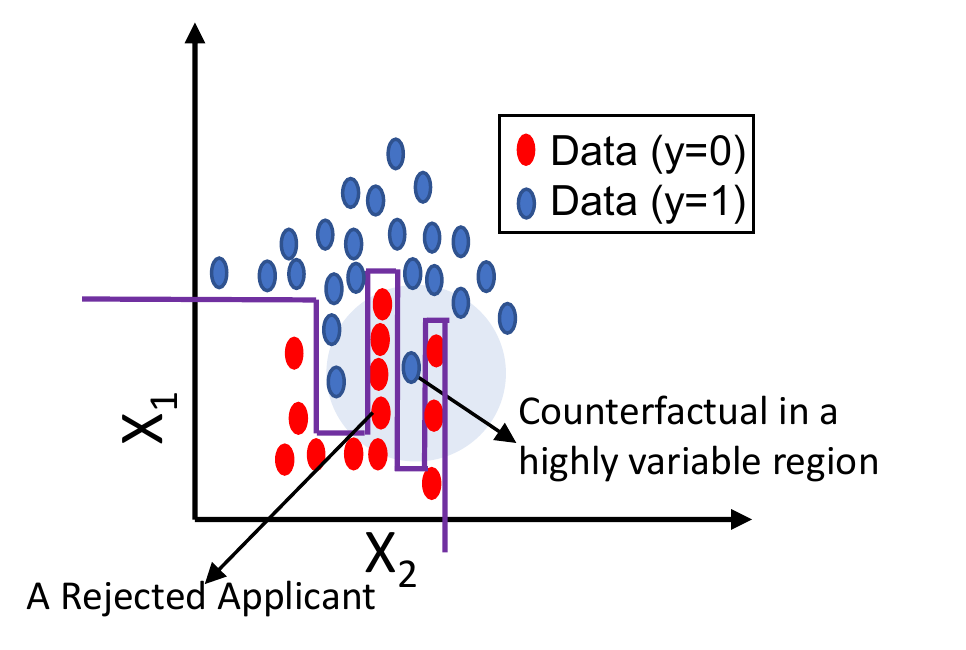}}
\caption{Counterfactual lies in a  highly variable region.}\label{fig:property3}
\end{subfigure}
\caption{Motivation for desirable properties.}
\end{figure}

However, having high $M(x)$ may not be the only property to ensure robustness, particularly in tree-based ensembles. This is because \emph{tree-based models do not have a smooth and continuous output function}. For instance, there may exist points $x \in \mathbb{R}^d$ with very high output value $M(x)$ but several points in its neighborhood have a low output value (not smooth). This issue is illustrated in Figure~\ref{fig:property2}. There may be points with high $M(x)$ that are quite close to the decision boundary, and thus more vulnerable to being invalidated with model changes. 

As a safeguard against such a possibility, we introduce our next desirable property.

\begin{propty}
An $x \in \mathbb{R}^d$ is less likely to be invalidated due to model changes if several points close to $x$ (denoted by $x'$) have a high value of $M(x')$.
\label{propty:high_confidence_mean}
\end{propty}

We also note that a counterfactual may be more likely to be invalidated if it lies in a highly variable region of the model output function $M(x)$. This is because the confidence of the model predictions in that region might be less reliable. This issue is illustrated in Figure~\ref{fig:property3}. One resolution to capturing the variability of a model output is to examine its derivative. However, because \emph{tree-based ensembles are not differentiable}, we instead examine the standard deviation of the model output around $x$ as a representative of its variability.

\begin{propty}
An $x \in \mathbb{R}^d$ is less likely to be invalidated due to model changes if the model output values around $x$ have low variability (standard deviation).
\label{propty:variability}
\end{propty}

\subsection{Proposed Quantification of Robustness to Possible Model Changes:\\ Counterfactual Stability}

Our properties lead us to introduce a novel metric -- that we call counterfactual stability -- that attempts to quantify the robustness of a counterfactual $x \in \mathbb{R}^d$ to possible changes in the model (irrespective of whether $x$ is in the data manifold).

\begin{defn}[Counterfactual Stability] The stability of a counterfactual $x\in \mathbb{R}^d$ is defined as follows: \begin{align}&R_{K,\sigma^2}(x,M)=\frac{1}{K}\sum_{x' \in N_x}M(x')  - \sqrt{\frac{1}{K}\sum_{x' \in N_x}\left(M(x') - \frac{1}{K}\sum_{x' \in N_x}M(x')\right)^2} 
\end{align}where $N_x$ is a set of $K$ points in $\mathbb{R}^d$ drawn from the distribution $\mathcal{N}(x,\sigma^2\mathrm{I}_{d})$ where $\mathrm{I}_{d}$ is the identity matrix.
\label{defn:stability}
\end{defn}

This metric of counterfactual stability is aligned with our desirable properties. Given a point $x\in \mathbb{R}^d$, it generates a set of $K$ points centered around $x$. The first term $\frac{1}{K}\sum_{x' \in N_x}M(x')$ is expected to be high if the model output value $M(x)$ is high for $x$ (Property~\ref{propty:high_confidence}) as well as several points close to $x$ (Property~\ref{propty:high_confidence_mean}). However, we note that the mean value of $M(x)$ around a point $x \in \mathbb{R}^d$ may not always capture the variability in that region. For instance, a combination of very high and very low values can also produce a reasonable mean value. Thus, we also incorporate a second term, i.e., the standard deviation $\sqrt{\frac{1}{K}\sum_{x' \in N_x}\left(M(x') - \frac{1}{K}\sum_{x' \in N_x}M(x')\right)^2} $ which captures the variability of the model output values in a region around $x$ (recall Property~\ref{propty:variability}). 

We also note that the variability term (standard deviation) in Definition~\ref{defn:stability} is useful only given the first term (mean) as well. This is because even points on the other side of the decision boundary (i.e., $M(x')< 0.5$) can have high or low variance. We include the histogram of $M(x)$, $\frac{1}{K}\sum_{x' \in N_x}M(x')$, and $R_{K,\sigma^2}(x,M)$ in the Appendix for further insights.

Next, we discuss how our proposed metric can be used to test if a counterfactual is \emph{stable}.

\begin{defn}[Counterfactual Stability Test] \label{defn:stability_test} A counterfactual $x\in \mathbb{R}^d$ satisfies the counterfactual stability test if: 
\begin{equation}R_{K,\sigma^2}(x,M)\geq \tau.
\end{equation}
\end{defn}

\begin{rem}[Discussion on Data Manifold] 
\label{rem:data_manifold}Our definition of counterfactual stability holds for all points $x\in \mathbb{R}^d$ and is not necessarily restricted to points that lie on the data manifold, e.g., $x \in \mathcal{S}$. This is because there might be points or regions outside the data manifold that could also be robust to model changes. E.g., assume a loan applicant who is exceptionally good at literally everything. Such an applicant might not lie on the data manifold, but it is expected that most models would accept such a data point even after retraining. We note however that recent work~\cite{pawelczyk2020counterfactual} demonstrate that data-support counterfactuals are more robust that sparse counterfactuals, an aspect that we discuss further in Section~\ref{subsec:robustness_guarantee} which also motivates our definition of conservative counterfactual. 
\end{rem}

\subsection{Concept of Conservative Counterfactuals}
\label{subsec:conservative_counterfactuals}
Here, we introduce the concept of \emph{Conservative Counterfactuals} which allows us to use our counterfactual stability test to generate stable counterfactuals from the dataset.

\begin{defn}[Conservative Counterfactual $\mathcal{C}^{(\tau)}_{p,\mathcal{S}}(x,M)$] Given a data point $x \in \mathcal{S}$ such that $M(x)\leq 0.5$, a conservative counterfactual $\mathcal{C}^{(\tau)}_{p,\mathcal{S}}(x,M)$ is defined as a data point $x' \in \mathcal{S}$ such that $M(x')>0.5$ and $R_{K,\sigma^2}(x',M)\geq\tau$, that also minimizes the $l_p$ norm $||x-x'||_p$, i.e.,
\begin{align}
&\mathcal{C}^{(\tau)}_{p,\mathcal{S}}(x,M)=\arg \min_{x'\in \mathcal{S}} ||x-x'||_p \nonumber \\
&\text{such that } M(x')>0.5 \text{ and } R_{K,\sigma^2}(x',M)\geq \tau.
\end{align}  
\end{defn}

\begin{rem}[Existence] Higher $\tau$ leads to better robustness. However, a conservative counterfactual may or may not exist depending on how high the threshold $\tau$ is.  When $\tau$ is very low, the conservative counterfactuals become the closest data-support counterfactuals.
\end{rem}


\begin{figure*}
    \centering
    \includegraphics[height=4cm]{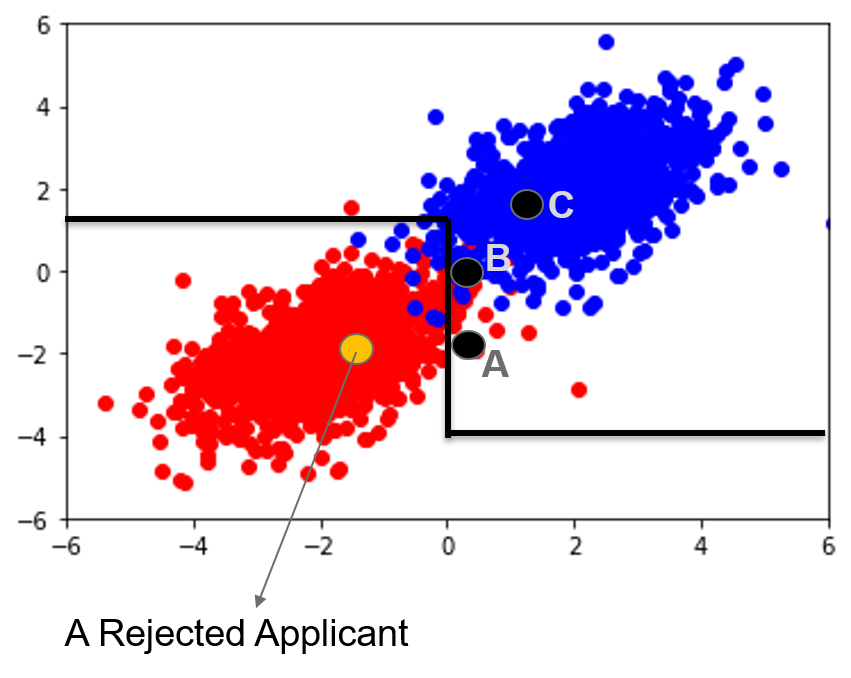}
    \includegraphics[height=4cm]{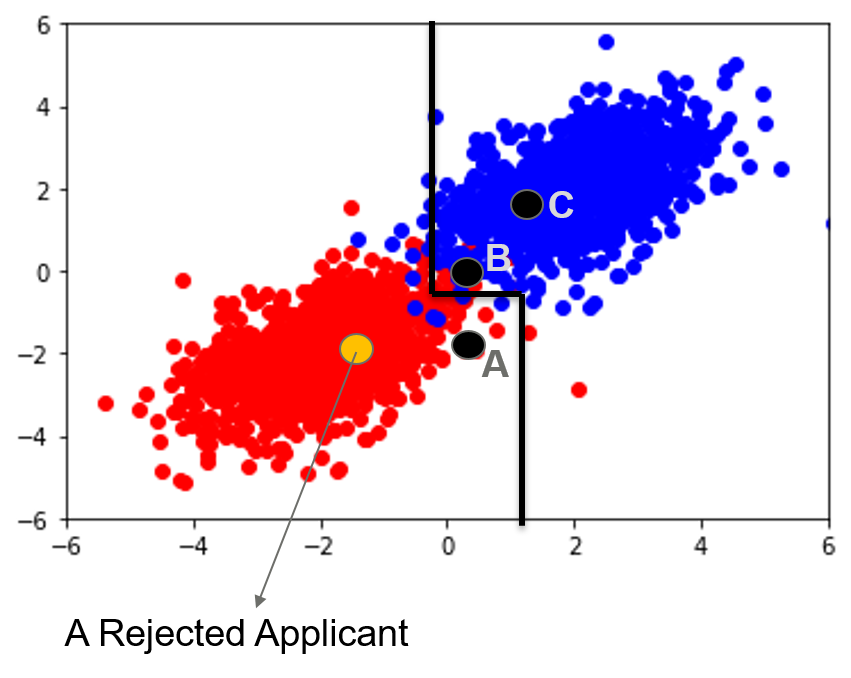}
    \includegraphics[height=4cm]{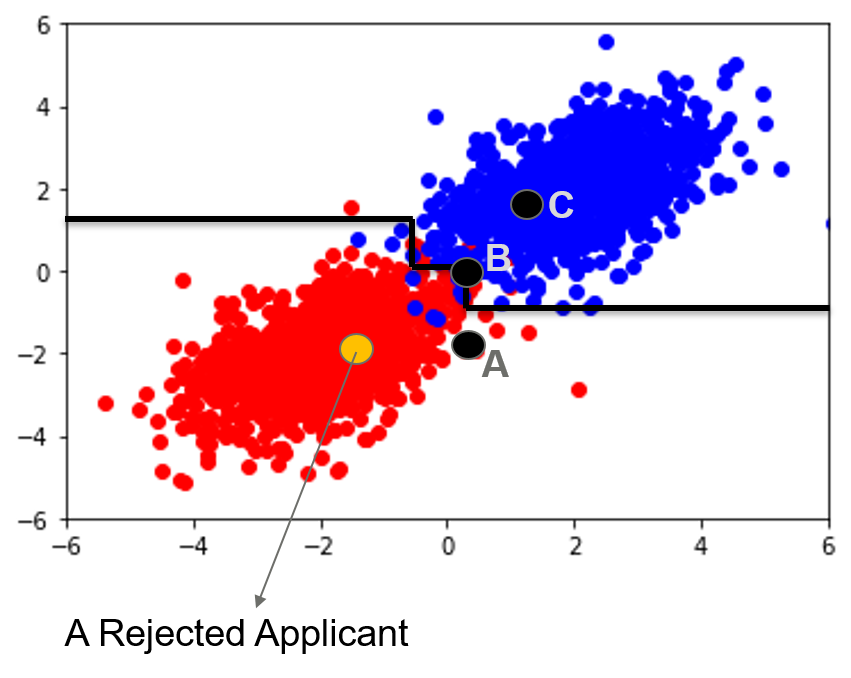}
    \caption{Thought experiment to understand how a conservative counterfactual is more robust than typical closest counterfactuals or closest-data-support counterfactuals: The data $\mathcal{S}$ is drawn from the following distribution: $p(x|y=1)\sim \mathcal{N}(\mu,\Sigma)$ and $p(x|y=0)\sim \mathcal{N}(-\mu,\Sigma)$. The first model denotes the original model $M(x)$ while the next two models denote possible models obtained after retraining on the same data with almost similar accuracy (performance) on the given dataset. Given a rejected applicant, we have $A$, $B$, and $C$ as three possible counterfactuals. Counterfactual $A$ is the closest counterfactual: it may not lie on the data manifold. Counterfactual $B$ is the closest-data-support counterfactual, i.e., the nearest neighbor on the other side of the decision boundary. The second figure demonstrates that data-support counterfactuals are more robust than closest (or, sparse) counterfactuals. However, lying on the data manifold is not always enough for robustness (third figure), e.g., $B$ happens to be quite close to the boundary. Here, $C$ is the conservative counterfactual that not only lies on the data manifold but also well within the decision boundary.}
    \label{fig:gaussian}
\end{figure*}

\subsection{Theoretical Robustness Guarantee of Conservative Counterfactuals}
\label{subsec:robustness_guarantee}
Here, we derive theoretical guarantees on the robustness of conservative counterfactuals (see Theorem~\ref{thm:guarantee} for our main result). Before stating our result, we introduce two assumptions over the randomness of the new model $M_{new}$.

\begin{assm}[Goodness of Metric] For any data point $x \in \mathbb{R}^d$, let $M_{new}(x)$ be a random variable taking different values due to model changes. 
We assume that the expected value $E[M_{new}(x)]>R_{K,\sigma^2}(x,M)$.
\label{assm:1}
\end{assm}

\begin{assm}[Goodness of Data Manifold]
The standard deviation of $M_{new}(x)$ is $V_x$ which depends on $x$. When $x \in \mathcal{S}$, we have $V_x \leq V$ for a small constant $V$. \label{assm:2}
\end{assm}

The rationale for Assumption~\ref{assm:2} is built on evidence from recent work~\cite{pawelczyk2020counterfactual} that demonstrate that data-support counterfactuals are more robust than closest or sparsest counterfactuals. When a model is retrained on same or similar data, the decisions of a model are less likely to change for points that lie on the data manifold as compared to points that may not lie on the data manifold (illustrated in Figure.~\ref{fig:gaussian}).

\begin{rem}[One-Way Implication]
While we assume that the new model outputs for points in the dataset $\mathcal{S}$ have low standard deviation, we do not necessarily assume that points outside the dataset $\mathcal{S}$ would always have high standard deviation. This is because there can potentially be regions outside the data manifold that also have low $V_x$, and are also robust to model changes (recall Remark~\ref{rem:data_manifold}).
\end{rem}

One popular assumption in existing literature to quantify small model changes is to assume that the model changes are bounded in the parameter space, i.e., $|\text{Parameters}(M)-\text{Parameters}(M_{new})|\leq \Delta$ where $\text{Parameters}(M)$ denote the parameters of the model $M$, e.g., weights of a neural network. However, this might not be a good assumption for tree-based ensembles. This is because tree-based ensembles can often change a lot in the parameter space while actually causing very little difference with respect to the actual decisions on the dataset $\mathcal{S}$ (see Figure~\ref{fig:gaussian}). 

Closely connected to model change is the idea of Rashomon models~\cite{pawelczyk2020counterfactual,marx2020predictive} which suggests that there can be models that are very different from each other but have almost similar performance on the same data, e.g., $\sum_{x \in \mathcal{S}} |D(x)-D_{new}(x)|\leq \Delta$. Thus, Assumption~\ref{assm:2} might be better suited for tree-based ensembles over boundedness in the parameter space.

Now, we provide our main result: a robustness guarantee on conservative counterfactuals based on these assumptions.







\begin{thm}[Robustness Guarantee for Conservative Counterfactuals] Suppose Assumptions~\ref{assm:1} and \ref{assm:2} hold, and $\tau>0.5$. Then, for any conservative counterfactual $x' \in \mathcal{C}^{(\tau)}_{p,\mathcal{S}}(x,M)$, the following holds:
\begin{equation}
    \Pr(M_{new}(x')< 0.5) \leq \frac{V^2}{V^2 + (\tau-0.5)^2}.
\end{equation}
\label{thm:guarantee}
\end{thm}

The result essentially says that the probability of invalidation by the new model ($\Pr(M_{new}(x')< 0.5)$) is strictly upper-bounded for conservative counterfactuals. A smaller variability $V$ makes this bound smaller. 

The conservative counterfactuals (henceforth denoted by CCF) already serve as good candidates for robust counterfactuals. They are also expected to be realistic with high LOF because they lie in the dataset $\mathcal{S}$. However, because they only search for counterfactuals on the dataset $\mathcal{S}$, they may not always be optimal in terms of the distance between the original data point and its counterfactual (not so close). This leads us to now propose a novel algorithm that leverages conservative counterfactuals (CCF) and counterfactual stability test to find robust counterfactuals that meet all our requirements (close, robust, realistic). 
\subsection{Proposed Algorithm to Generate Robust Counterfactuals in Practice: RobX}

In this section, we discuss our proposed algorithm -- that we call RobX -- that generates robust counterfactuals that meets our requirements (see Algorithm~\ref{alg:example}). 

Our proposed algorithm RobX can be applied on top of any preferred base method of counterfactual generation, irrespective of whether the counterfactual lies in the dataset $\mathcal{S}$. RobX checks if the generated counterfactual satisfies the counterfactual \emph{stability test} (recall Definition~\ref{defn:stability_test}): if the test is not satisfied, the algorithm iteratively refines the obtained counterfactual and keeps moving it towards the conservative counterfactual until a \emph{stable} counterfactual is found that satisfies the test. 

One might wonder if moving a counterfactual towards the conservative counterfactual can cause it to pass through undesired regions of the model output where $M(x)<0.5$, thus making it more vulnerable to invalidation. We note that, while this concern is reasonable, the counterfactual \emph{stability test} at each step ensures that such points are not selected. We further address this concern as follows: (i) consider a diverse set of conservative counterfactuals (e.g., first $c$ nearest neighbors that satisfy the stability test where $c>1$); (ii) iteratively move towards each one of them until a \emph{stable} counterfactual is found for all $c$ cases; (iii) pick the best of these $c$ \emph{stable} counterfactuals, e.g., one with the lowest $L_1$ or $L_2$ cost as desired. 

We also observe that this approach of moving a counterfactual towards a conservative counterfactual improves its LOF, making it more realistic.

\begin{algorithm}[t]
   \caption{RobX: Generating Robust Counterfactual Explanations for Tree-Based Ensembles}
   \label{alg:example}
\begin{algorithmic}
   \STATE {\bfseries Input:} Model $M(\cdot)$, Dataset $\mathcal{S}$, Datapoint $x$ such that $M(x) \leq 0.5$, Algorithm parameters $(p, K, \sigma^2, \tau, \alpha, c)$
   \STATE{Step 1: Generate counterfactual $x'$ for $x$ using any existing technique for tree-based ensembles}
   \STATE{Step 2: Perform counterfactual stability test on $x'$: Check if $R_{K,\sigma^2}(x',M)\geq \tau$ where $N_x$ is a set of $K$ points drawn from the distribution $\mathcal{N}(x,\sigma^2)$.}
   \IF{counterfactual stability test is satisfied:}
   \STATE{Output $x'$ and exit}
   \ELSE
   \STATE{Generate $c$ conservative counterfactuals $\{x_1,\ldots,x_c\}$ which are $c$ nearest neighbors of $x'$ in the dataset $\mathcal{S}$ that pass the stability test: $R_{K,\sigma^2}(x_i,M)\geq \tau$}
   \STATE{Initialize placeholders for $c$ counterfactuals $\{x'_1,\ldots,x'_c\}$ with each $x'_i=x'$}
   \FOR {$i=1 \text{ to } c$}
   \REPEAT 
   \STATE{Update: $x'_i=\alpha x_i + (1-\alpha)x'_i$}
   \STATE{Perform  counterfactual stability test on $x'_i$:\\ \hspace{1cm} $R_{K,\sigma^2}(x'_i,M)\geq \tau$}
   \UNTIL counterfactual stability test on $x'_i$ is satisfied
    \ENDFOR
      \ENDIF
    \STATE{Output $x^*=\arg \min_{x'_i \in \{x'_1,x'_2,\ldots,x_c'\}} ||x-x'_i||_p$ and exit}
\end{algorithmic}
\end{algorithm}

\section{Experiments}

Here, we present our experimental results on benchmark datasets, namely, German Credit~\cite{UCI} and HELOC~\cite{Fico_web_2018}. 

For simplicity, we normalize the features to lie between $[0,1]$. We consider XGBoost models after selecting hyperparameters from a grid search (details in Appendix). For each of these datasets, we set aside 30\% of the dataset for testing, and use the remaining 70\% for training (in different configurations as discussed here). 

We consider the following types of model change scenarios:
\begin{itemize}[leftmargin=*, topsep=0pt, itemsep=0pt]
    \item Minor changes: (i) Train a model on the training dataset and retrain new models after dropping very few data points ($1$ for German Credit, $10$ for HELOC), keeping hyperparameters constant. (ii) Train a model on the training dataset and retrain new models, changing one hyperparameter, e.g., \texttt{max\_depth} or \texttt{n\_estimators}. The results for this configuration is in the Appendix.
    \item Moderate changes: Train a model on half of the training dataset and retrain new models on the other half, keeping hyperparameters mostly constant, varying either \texttt{max\_depth} or \texttt{n\_estimators}. The results for this configuration is in Table~\ref{table:performance}.
\end{itemize}

\begin{table}[t]
\caption{Performance on HELOC and German Credit dataset.}
\label{table:performance}
\begin{center}
\begin{small}
\begin{sc}
\begin{tabular}{c|p{0.9cm}p{0.9cm}p{0.9cm}|p{0.9cm}p{0.9cm}p{0.9cm}}
\toprule
\textbf{HELOC}& \multicolumn{3}{c|}{$L_1$ Based} & \multicolumn{3}{c}{$L_2$ Based}\\
\midrule
Method &  Cost & Val. & LOF & Cost & Val. & LOF\\
\midrule
CCF     & 1.89 & 100\% & 0.81 & 0.65& 99.9\%& 0.75\\
\midrule
FT    & 0.19 & 18.7\%& 0.40  & 0.16 & 15.6\%&  0.48\\
+RobX & 1.55 & 100\% &   0.92 & 0.55  & 99.9\% & 0.84    \\
\midrule
FOCUS    & 0.21 & 29.5\% & 0.36 & 0.17 & 33.0\% & 0.63\\
+RobX & 1.52& 100\% & 0.91 & 0.61& 99.8\% & 0.72\\
\midrule
FACE  & 2.86 & 89.4\%&      0.68 &  1.19 & 97.3\% & 0.50  \\
+RobX & 2.30 & 100\% & 0.78 &  0.95& 100\% & 0.65\\
\midrule
NN   & 0.96 & 35.1\% & 0.81 & 0.34 & 39.0\%& 0.69 \\
+RobX  & 1.61 & 100\% & 0.93 & 0.56 & 100\% & 0.85\\
\bottomrule
\end{tabular}
\end{sc}
\end{small}
\begin{small}
\begin{sc}
\begin{tabular}{c|p{0.9cm}p{0.9cm}p{0.9cm}|p{0.9cm}p{0.9cm}p{0.9cm}}
\toprule
\textbf{German}& \multicolumn{3}{c|}{$L_1$ Based} & \multicolumn{3}{c}{$L_2$ Based}\\
\midrule
Method & Cost & Val. & LOF & Cost & Val. & LOF\\
\midrule
CCF     & 2.92 & 100\% & 0.85 & 1.21 & 100\% & 0.94\\
\midrule
FT    & 0.13 & 55.7\% & 0.93 & 0.11 & 59.2\% & 0.94 \\
+RobX & 2.17  & 92.6\% &  1.0  & 0.95 &  91.1\% & 0.94  \\
\midrule
FOCUS    & 0.37  & 65.7\% & 0.93 & 0.24 & 65.3\% & 0.93\\
+RobX & 2.18 & 96.5\% & 1.0 & 1.05 & 100\% & 1.0\\
\midrule
FACE  & 2.65 & 84.5\% &  0.57 & 1.30 & 87.6\% &   0.76 \\
+RobX & 2.29  & 97.1\% & 1.0 & 1.05 & 96.1\% & 0.94\\
\midrule
NN   & 0.76 & 65.9\% & 1.0 & 0.48 & 60.7\% & 1.0\\
+RobX  & 2.21 & 97.7\% & 1.0 & 0.97 & 91.7\% & 0.93 \\
\bottomrule
\end{tabular}
\end{sc}
\end{small}
\end{center}
\vskip -0.1in
\end{table}

For each case, we first generate counterfactuals for the original model using the following base methods:
\begin{itemize}[leftmargin=*, topsep=0pt, itemsep=0pt]
    \item Feature Tweaking (FT)~\cite{tolomei2017interpretable} is a popular counterfactual generation technique for tree-based ensembles that finds ``closest'' counterfactuals ($L_1$ or $L_2$ cost), not necessarily on the data manifold. The algorithm searches for all possible paths (tweaks) in each tree that can change the final outcome of the model.
    \item  FOCUS~\cite{lucic2019focus} is another popular technique that approximates the tree-based models with \texttt{sigmoid} functions, and finds closest counterfactuals (not necessarily on the data manifold) by solving an optimization.
    \item  FACE~\cite{poyiadzi2020face} attempts to find counterfactuals that are not only close ($L_1$ or $L_2$ cost), but also (i) lie on the data manifold; and (ii) are connected to the original data point via a path on a connectivity graph on the dataset $\mathcal{S}$. Such a graph is generated from the given dataset $\mathcal{S}$  by connecting every two points that are reasonably close to each other, so that one can be ``attained'' from the other. 
    \item Nearest Neighbor (NN)~\cite{albini2021counterfactual} attempts to find counterfactuals that are essentially the nearest neighbors ($L_1$ or $L_2$ cost) to the original data points with respect to the dataset $\mathcal{S}$ that lie on the other side of the decision boundary (recall Definition~\ref{defn:data-support-CF}).
\end{itemize}

We compare these base methods with: (i) Our proposed Conservative Counterfactuals (CCF) approach; and (ii) Our proposed RobX applied on top of these base methods.



\begin{rem}
We note that there are several techniques for generating counterfactual explanations (see \cite{verma2020counterfactual} for a survey); however only some of them apply to tree-based models. Several techniques are also broadly similar to each other in spirit. We believe our choice of these four base methods to be quite a diverse representation of the existing approaches, namely, search-based closest counterfactual (FT), optimization-based closest counterfactual (FOCUS),  graph-based data-support counterfactual (FACE), and closest-data-support counterfactual (NN). We note that another alternative perspective is a causal approach~\cite{konig2021causal}, that often requires knowledge of causal structure, which is outside the scope of this work.
\end{rem}

Our empirical performance metrics of interest are:
\begin{itemize}[leftmargin=*, topsep=0pt, itemsep=0pt]
    \item \textbf{Cost ($L_1$ or $L_2$):} Average distance ($L_1$ or $L_2$) between the original point and its counterfactual.
    \item \textbf{Validity (\%):} Percentage of counterfactuals that still remain counterfactuals under the new model $M_{new}$.
    \item \textbf{LOF}: See Definition~\ref{defn:lof}; Implemented using \cite{scikit-lof} (+1 for inliers, -1 otherwise). A higher average is better.
\end{itemize}

\textbf{Hyperparameters:} For the choice of $K$ and $\sigma$, we refer to some guidelines in adversarial machine learning literature. Our metric of stability is loosely inspired from certifiable robustness in adversarial machine learning literature~\cite{cohen2019certified,raghunathan2018certified}, which uses the metric $\frac{1}{K}\sum_{x' \in N_x}I(M(x')>0.5)$. Here $I(.)$ is the indicator function. Our metric for counterfactual stability (in Definition~\ref{defn:stability}) has some key differences: (i) No indicator function; (ii) We leverage the standard deviation as well along with the mean.  Because the feature values are normalized, a fixed choice of $K=1000$ and $\sigma=0.1$ is used for all our experiments. 

The choice of threshold $\tau$ however, is quite critical, and depends on the dataset. As we increase $\tau$ for conservative counterfactuals, the validity improves but the $L_1$/$L_2$ cost also increases, until the validity almost saturates. If we increase $\tau$ beyond that, there are no more conservative counterfactuals found. In practice, one can examine the histogram of $R_{K,\sigma^2}(x',M)$ for $x \in \mathcal{S}$, and choose an appropriate quantile for that dataset as $\tau$ so that a reasonable fraction of points in $\mathcal{S}$ qualify to be conservative counterfactuals. But, \emph{the same quantile may not suffice for $\tau$ across different datasets.} One could also perform the following steps: (i) choose a small validation set; (ii) keep increasing $\tau$ from $0.5$ for CCF and plot the validity and $L1/L2$ cost; (iii) select a $\tau$ beyond which validity does not improve much and the $L1/L2$ cost is acceptable.

Next, we include the experimental results for moderate changes to the model in  Table~\ref{table:performance} for both HELOC and German Credit datasets. Additional results are provided in the Appendix.

\begin{table}
\caption{Performance of FOCUS on the model with a higher threshold, i.e., $M(x)> \gamma$ on HELOC and German Credit datasets.}
\label{table:threshold_focus}
\begin{center}
\begin{small}
\begin{sc}
\begin{tabular}{c|p{0.9cm}p{0.9cm}p{0.9cm}|p{0.9cm}p{0.9cm}p{0.9cm}}
\toprule
\textbf{HELOC}& \multicolumn{3}{c|}{$L_1$ Based} & \multicolumn{3}{c}{$L_2$ Based}\\
\midrule
Method &  Cost & Val. & LOF & Cost & Val. & LOF\\
\midrule
$\gamma{=}0.5$ & 0.21   & 29.5\% &  0.36  &  0.17 & 33.0\% &  0.63 \\
+RobX   & 1.52 & 100\% & 0.91 & 0.55 & 99.9\% & 0.84\\
\midrule
$\gamma{=}0.7$ & 0.59& 92.2\% & -0.01& 0.34 & 98.8\% & 0.30\\
 +RobX & 1.38 & 99.9\% & 0.70 & 0.60 & 99.8\% & 0.70 \\
\midrule
$\gamma{=}0.75$ & 1.13 & 98.9\% & -0.32& 0.44 & 99.9\% & 0.09\\
 +RobX  & 1.44 & 100\% & 0.06& 0.55 & 99.9\% & 0.51\\
\midrule
$\gamma{=}0.8$ & 2.11 & 100\% & -0.70& 0.60 & 100\% & -0.20\\
+RobX & 2.14 & 100\% & -0.66 & 0.62 & 100\% & -0.08\\
\bottomrule
\end{tabular}
\end{sc}
\end{small}
\begin{small}
\begin{sc}
\begin{tabular}{c|p{0.9cm}p{0.9cm}p{0.9cm}|p{0.9cm}p{0.9cm}p{0.9cm}}
\toprule
\textbf{German} & \multicolumn{3}{c|}{$L_1$ Based} & \multicolumn{3}{c}{$L_2$ Based}\\
\midrule
Method &  Cost & Val. & LOF & Cost & Val. & LOF\\
\midrule
$\gamma{=}0.5$ & 0.37   & 65.7\% &  0.93 & 0.24 & 65.3\% & 0.93 \\
 +RobX   & 2.18 & 96.5\% & 1.0 & 1.05 & 100\% & 1.0 \\
\midrule
$\gamma{=}0.8$ & 0.74& 83.9\% & 0.87 & 0.41& 92.1\% & 0.75\\
 +RobX & 2.20 & 98.0\% & 1.0 & 1.05 & 100\% & 1.0\\
\midrule
$\gamma{=}0.99$ & 2.57 & 97.7\% & -0.45 & 1.05 & 98.5\% & -0.33\\
 +RobX & 2.85 & 100\% & 0.03& 1.19 & 100\% & -0.27\\
\bottomrule
\end{tabular}
\end{sc}
\end{small}
\end{center}
\vskip -0.1in
\end{table}


\textbf{Observations:} The average cost ($L_1$ or $L_2$ cost) between the original data point and the counterfactual  increases only slightly for base methods such as FT, FOCUS, and NN (which find counterfactuals by explicitly minimizing this cost); however our counterfactuals are significantly more robust (in terms of validity) and realistic (in terms of LOF). Interestingly, for FACE (which finds counterfactuals on the data manifold that are connected via a path), our strategy is able to improve both robustness (validity) and cost ($L_1$ or $L_2$ cost), with almost similar LOF.

Another competing approach that we consider in this work (that has not been considered before) is to find counterfactuals using base methods but setting a higher threshold value for the model, i.e., $M(x) > \gamma$ where $\gamma$ is greater than $0.5$. Interestingly, we observe that this simple modification can also sometimes generate counterfactuals that are significantly robust; however this approach has several disadvantages: (i) It generates counterfactuals that are quite unrealistic, and thus have very poor LOF. (ii) The algorithm takes significantly longer to find counterfactuals as the threshold $\gamma$ is increased, and sometimes even returns a \texttt{nan} value because no counterfactual is found, e.g., if $\gamma=0.9$ and the model output $M(x)$ rarely takes such a high value (because the output range of tree-based ensembles) takes discrete values). Because of these disadvantages, we believe this technique might not be preferable to use standalone; however, it can be used as an alternate base method over which our technique might be applied when cost ($L_1$ or $L_2$) is a higher priority over LOF (see Table~\ref{table:threshold_focus}).




\textbf{Discussion and Future Work:} This work addresses the problem of finding robust counterfactuals for tree-based ensembles. It provides a novel metric to compute the stability of a counterfactual that can be representative of its robustness to possible model changes, as well as, a novel algorithm to find robust counterfactuals. Though not exactly comparable, but our cost and validity are in the same ballpark as that observed for these datasets in existing works~\cite{upadhyay2021towards,black2021consistent}, focusing on robust counterfactuals for linear models or neural networks (differentiable models). Our future work would include: (i) extending to causal approaches~\cite{konig2021causal}; and (ii) accounting for immutability or  differences among features, e.g., some features being more variable than others. 


\paragraph{Disclaimer}
This paper was prepared for informational purposes by
the Artificial Intelligence Research group of JPMorgan Chase \& Co. and its affiliates (``JP Morgan''),
and is not a product of the Research Department of JP Morgan.
JP Morgan makes no representation and warranty whatsoever and disclaims all liability,
for the completeness, accuracy or reliability of the information contained herein.
This document is not intended as investment research or investment advice, or a recommendation,
offer or solicitation for the purchase or sale of any security, financial instrument, financial product or service,
or to be used in any way for evaluating the merits of participating in any transaction,
and shall not constitute a solicitation under any jurisdiction or to any person,
if such solicitation under such jurisdiction or to such person would be unlawful.

\section*{Acknowledgements} We thank Emanuele Albini and Dan Ley for useful discussions.

\small{
\bibliography{references}
\bibliographystyle{IEEEtran}
}


\appendix

\section{Proof of Theorem 1}

The proof follows by demonstrating that another tree-based ensemble model exists that does not accept the generated set of counterfactuals. Let us denote this set as $\mathcal{CF}$.

There are various ways to construct such a model. A simple way could be to choose an identical tree structure but with the predictions flipped, i.e., $M_{new}(x)=1-M(x)$. This can be designed by altering the weights of the leaves.

For an $x \in \mathcal{CF}$ with $M(x)>0.5$, we would have $M_{new}(x) \leq 0.5$.

\section{Proof of Theorem 2}

The proof follows from Cantelli's inequality. The inequality states that, for $ \lambda >0$,
$$ \Pr(Z-\mathbb{E}[Z]\leq -\lambda )\leq {\frac {V_Z^{2}}{V_Z ^{2}+\lambda ^{2}}},$$
where $Z$ is a real-valued random variable,
$\Pr $ is the probability measure,
$ \mathbb{E}[Z]$ is the expected value of $Z$, and
$ V_Z^{2}$ is the variance of $Z$.

Here, let $Z=M_{new}(x')$ be a random variable that takes different values for different models $M_{new}$. Let $$\lambda=\mathbb{E}[Z]-0.5 \geq R(x',M,K,\sigma^2)-0.5> \tau-0.5 > 0.$$

Then, we have:
\begin{align}
    \Pr(Z\leq 0.5 ) 
    &= \Pr(Z -\mathbb{E}[Z] \leq 0.5-\mathbb{E}[Z] ) \nonumber \\
    & = \Pr(Z -\mathbb{E}[Z] \leq -\lambda ) \text{ where } \lambda=\mathbb{E}[Z]-0.5 \nonumber \\
    & \overset{(a)}{\leq} \frac{V_Z^2}{V_Z^2+\lambda^2}  \overset{(b)}{\leq} \frac{V_Z^2}{V_Z^2+(\tau-0.5)^2}  \overset{(c)}{\leq} \frac{V^2}{V^2+(\tau-0.5)^2}.
\end{align}
Here, (a) holds from Cantelli's inequality, (b) holds because $\lambda> \tau-0.5$ (from the conditions of the theorem), and (c) holds because the variance of $Z$ is bounded by $V^2$ from Assumption 2.

\section{Additional Details and Experiments}

Here, we include further details related to our experiments, as well as, some additional results.

\subsection{Datasets}

\begin{itemize}[topsep=0pt, itemsep=0pt, leftmargin=*]
    \item HELOC~\cite{Fico_web_2018}: This dataset has $10K$ data points, each with $23$ finance-related features. We drop the features \texttt{MSinceMostRecentDelq}, \texttt{MSinceMostRecentInqexcl7days}, and \texttt{NetFractionInstallBurden}. We also drop the data points with missing values. Our pre-processed dataset has $n=8291$ data points with $d=20$ features each.
    \item German Credit~\cite{UCI}: This dataset $1000$ data points, each with $20$ features. The features are a mix of numerical and categorical features. We only use the following $10$ features: \texttt{existingchecking}, \texttt{credithistory}, \texttt{creditamount}, \texttt{savings}, \texttt{employmentsince}, \texttt{otherdebtors}, \texttt{property}, \texttt{housing}, \texttt{existingcredits}, and \texttt{job}. Among these features, we convert the categorical features into appropriate numeric values. E.g., \texttt{existingchecking} originally has four categorical values: A11 if ... < 0 DM, A12 if 0 <= ... < 200 DM, A13 if ... >= 200 DM, and A14 if no checking account. We convert them into numerical values as follows: $0$  for A14, $1$  for A11, $2$ for A12, and $3$ for A13. Our pre-processed dataset has $n=1000$ data points with $d=10$ features each.
\end{itemize}

All features are normalized to lie between $[0,1]$.

\subsection{Experimental Results Under Minor Changes to the Model}

For each of the datasets, we perform a $30/70$ test-train split. We train an XGBoost Model after tuning the hyperparameters using the \texttt{hyperopt} package. 

The observed accuracy for the two datasets are:  74\% (HELOC) and 73\% (German Credit).

For RobX, we choose $K=1000$, $\sigma=0.1$, and $\tau$ is chosen based on the histogram of $R_{K,\sigma^2}(x,M)$ for each dataset. For HELOC, $\tau=0.65$, and for German Credit, $\tau=0.93$.

We first present our experimental results after minor changes to the model.\\

(i) Train a model ($M(x)$) on the training dataset and retrain new models ($M_{new}(x)$) after dropping a small percentage of data points ($1$ for German Credit, $10$ for HELOC), keeping hyperparameters fairly constant. We experiment with $20$ different new models and report the average values in Tables~\ref{table:heloc_minor} and \ref{table:german_minor}.

\begin{table}[!htbp]
\caption{Performance on HELOC dataset minimizing for $L_1$ and $L_2$ cost.}
\label{table:heloc_minor}
\begin{center}
\begin{small}
\begin{sc}
\begin{tabular}{lccc}
\toprule
Method & $L_1$ Cost & Validity & LOF \\
\midrule
CCF     & 1.83 & 100\% & 0.89\\
\midrule
FT    & 0.19 & 71.1\%& 0.25  \\
FT +RobX & 1.51 & 100\% & 0.96\\
\midrule
FOCUS  & 0.22     & 77.8\% & 0.26 \\
FOCUS +RobX & 1.49 & 100\% & 0.94 \\
\midrule
FACE  & 2.95 & 98.9\%&   0.70     \\
FACE +RobX & 2.26  & 100\%  & 0.91\\
\midrule
NN   & 1.01 & 85.3\% & 0.75 \\
NN +RobX  & 1.56 & 100\% & 0.96 \\
\bottomrule
\end{tabular}
\end{sc}
\end{small}
\begin{small}
\begin{sc}
\begin{tabular}{lcccr}
\toprule
Method & $L_2$ Cost & Validity & LOF \\
\midrule
CCF     & 0.62 & 100\% & 0.83\\
\midrule
FT    & 0.16 & 68.3\% & 0.40  \\
FT +RobX & 0.54 & 100\% & 0.93 \\
\midrule
FOCUS    & 0.16 & 57.1\% & 0.57 \\
FOCUS +RobX & 0.59 & 100\% & 0.86\\
\midrule
FACE  & 1.20 & 99.6\% &   0.58     \\
FACE +RobX & 0.89  & 100\%  & 0.74 \\
\midrule
NN   & 0.35 &  84.0\%& 0.75 \\
NN +RobX  & 0.55 &  100\%& 0.95 \\
\bottomrule
\end{tabular}
\end{sc}
\end{small}
\end{center}
\vskip -0.1in
\end{table}
\begin{table}[!htbp]
\caption{Performance on German Credit dataset minimizing for $L_1$ and $L_2$ cost.}
\label{table:german_minor}
\begin{center}
\begin{small}
\begin{sc}
\begin{tabular}{lccc}
\toprule
Method & $L_1$ Cost & Validity & LOF \\
\midrule
CCF     & 3.05 & 100\% & 1.0\\
\midrule
FT    & 0.08 & 72.9\%& 0.65  \\
FT +RobX & 2.70 & 99.7\% & 1.0\\
\midrule
FOCUS    & 0.12 & 72.8\% & 0.71 \\
FOCUS +RobX & 2.71 & 100\% & 1.0\\
\midrule
FACE  & 2.67 & 92.5\% &  0.94      \\
FACE +RobX & 2.70 & 99.1\% & 1.0\\
\midrule
NN   & 0.80 & 84.4\%& 0.94 \\
NN +RobX  & 2.71 & 100\% & 1.0\\
\bottomrule
\end{tabular}
\end{sc}
\end{small}
\begin{small}
\begin{sc}
\begin{tabular}{lcccr}
\toprule
Method & $L_2$ Cost & Validity & LOF \\
\midrule
CCF     & 1.42  & 100\%& 1.0\\
\midrule
FT    & 0.08 & 68.7\%& 0.65  \\
FT +RobX & 1.27 & 100\% & 1.0 \\
\midrule
FOCUS    & 0.11 & 70.1\% & 0.82 \\
FOCUS +RobX & 1.32 & 100\% & 1.0\\
\midrule
FACE  & 1.25 & 95.0\% & 0.77       \\
FACE +RobX & 1.28 & 100\% & 1.0\\
\midrule
NN   & 0.49 & 79.1\%& 0.88  \\
NN +RobX  & 1.30 & 100\% & 1.0\\
\bottomrule
\end{tabular}
\end{sc}
\end{small}
\end{center}
\vskip -0.1in
\end{table}
(ii) Train a model $M(x)$ on the training dataset and retrain new models, changing one hyperparameter, e.g., \texttt{max\_depth} or \texttt{n\_estimators}. We experiment with 20 different new models and report the average values in Tables~\ref{table:heloc_minor2} and \ref{table:german_minor2}.

\begin{table}[H]
\caption{Performance on HELOC dataset minimizing for $L_1$ and $L_2$ cost.}
\label{table:heloc_minor2}
\vskip 0.1in
\begin{center}
\begin{small}
\begin{sc}
\begin{tabular}{lccc}
\toprule
Method & $L_1$ Cost & Validity & LOF \\
\midrule
CCF     & 1.83 & 98.9\% & 0.89\\
\midrule
FT    & 0.19 &49.9\% & 0.25  \\
FT +RobX &1.51  & 99.4\% & 0.96\\
\midrule
FOCUS  & 0.22    & 37.1\% & 0.26 \\
FOCUS +RobX &1.49 & 99.2\% & 0.94 \\
\midrule
FACE  & 2.67 &89.7\% &  0.94      \\
FACE +RobX & 2.70 & 99.7\% & 1.0\\
\midrule
NN   & 1.01 & 71.3\%& 0.75 \\
NN +RobX & 1.56  &99.6\% & 0.96\\
\bottomrule
\end{tabular}
\end{sc}
\end{small}
\begin{small}
\begin{sc}
\begin{tabular}{lcccr}
\toprule
Method & $L_2$ Cost & Validity & LOF \\
\midrule
CCF     & 0.62  & 98.7\% & 0.82\\
\midrule
FT    &0.16  & 49.4\%&  0.40 \\
FT +RobX &0.54  & 99.3\% & 0.92 \\
\midrule
FOCUS    & 0.16 & 50.3\% & 0.57 \\
FOCUS +RobX & 0.59 & 99.9\% & 0.86\\
\midrule
FACE  & 1.20 & 89.5\% &  0.59      \\
FACE +RobX & 0.89 & 99.9\% & 0.75\\
\midrule
NN   & 0.35 & 71.0\% & 0.74 \\
NN +RobX  & 0.55 &  99.6\% & 0.95\\
\bottomrule
\end{tabular}
\end{sc}
\end{small}
\end{center}
\vskip -0.1in
\end{table}
\begin{table}[H]
\caption{Performance on German Credit dataset minimizing for $L_1$ and $L_2$ cost.}
\label{table:german_minor2}
\vskip 0.1in
\begin{center}
\begin{small}
\begin{sc}
\begin{tabular}{lccc}
\toprule
Method & $L_1$ Cost & Validity & LOF \\
\midrule
CCF     & 3.05 & 99.9\% & 1.0\\
\midrule
FT    & 0.08  & 56.4\%& 0.65  \\
FT +RobX & 2.70  & 99.9\% & 1.0 \\
\midrule
FOCUS    & 0.12 & 53.7\% & 0.71 \\
FOCUS +RobX & 2.71 & 99.7\% & 1.0\\
\midrule
FACE  & 2.62 & 88.8\%&  0.82      \\
FACE +RobX & 2.72 & 99.7\% & 1.0\\
\midrule
NN   & 0.80 & 84.4\% & 0.94 \\
NN +RobX  & 2.71& 99.7\%& 1.0 \\
\bottomrule
\end{tabular}
\end{sc}
\end{small}
\begin{small}
\begin{sc}
\begin{tabular}{lcccr}
\toprule
Method & $L_2$ Cost & Validity & LOF \\
\midrule
CCF     & 1.36 & 97.4\% & 1.0 \\
\midrule
FT    & 0.08 & 53.4 & 0.65  \\
FT +RobX & 1.17 & 98.6  & 1.0 \\
\midrule
FOCUS    & 0.11 & 53.2\% & 0.82 \\
FOCUS +RobX &1.2 & 100\% &1.0 \\
\midrule
FACE  & 1.25 & 88.7\%&   0.77     \\
FACE +RobX & 1.18 & 98.4\% & 1.0\\
\midrule
NN   & 0.49 & 79.0\% & 0.88  \\
NN +RobX  & 1.18& 99.0\% & 0.94\\
\bottomrule
\end{tabular}
\end{sc}
\end{small}
\end{center}
\vskip -0.1in
\end{table}



\subsection{Histograms}
\begin{figure}[!htbp]
\centering
\includegraphics[height=3cm]{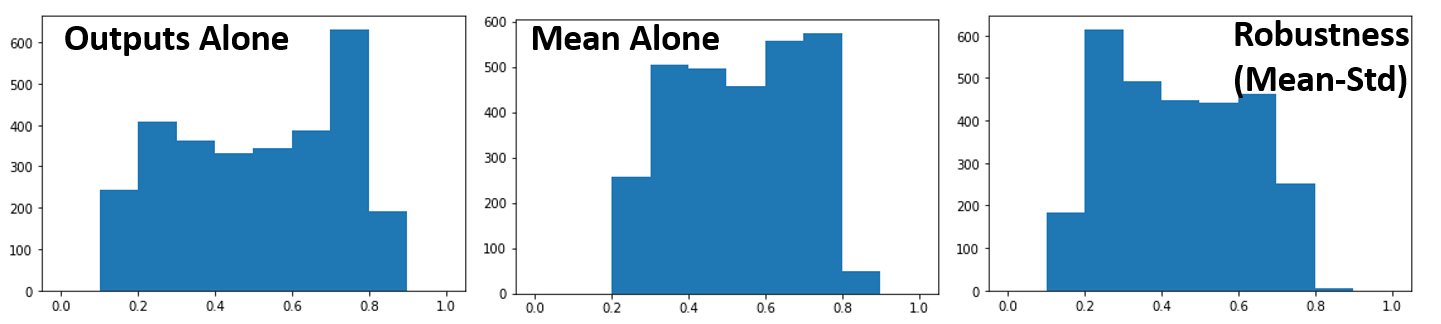}
\caption{Histograms to visualize the proposed robustness metric. \label{fig:ablation}}
\end{figure}

Here, we include the following histograms for the HELOC dataset for further insights (see Figure~\ref{fig:ablation}):
\begin{enumerate}[itemsep=0pt, topsep=0pt, leftmargin=*]
\item Model outputs alone, i.e., $M(x)$.
\item Mean of the model outputs in a neighborhood, i.e., $\frac{1}{K}\sum_{x' \in N_x}M(x')$.
\item Our robustness metric which includes the mean of the model outputs in a neighborhood minus their standard deviation, i.e., $R_{K,\sigma^2}(x,M)=\frac{1}{K}\sum_{x' \in N_x}M(x')- \sqrt{\frac{1}{K}\sum_{x' \in N_x}\left(M(x') - \frac{1}{K}\sum_{x' \in N_x}M(x')\right)^2}$.
\end{enumerate}

\subsection{Experiments Under Major Changes to the Model}

These experimental results have already been included in the main paper in Section 4. Here we include some additional details. For each of the datasets, we first perform a $30/70$ test-train split, and set the test data aside.

On the training data, we again perform a $50/50$ split. We train the original model $M(x)$ on one of these splits, and the new model $M_{new}(x)$ on the other. For $M(x)$, we train an XGBoost Model after tuning the hyperparameters using the \texttt{hyperopt} package. For $M_{new}(x)$, we keep the hyperparameters mostly constant, varying either \texttt{n\_estimators} or \texttt{max\_depth}.

The observed accuracy for the two datasets are:  73\% (HELOC) and 71\% (German Credit).

For RobX, we choose $K=1000$, $\sigma=0.1$, and $\tau$ is chosen based on the histogram of $R_{K,\sigma^2}(x,M)$ for each dataset. For HELOC, $\tau=0.65$, and for German Credit, $\tau=0.93$.
\subsection{Additional Experimental Results}

In our experiments so far, we normalize the features in the dataset to lie between $0$ and $1$ as is done in existing works (using \texttt{MaxMinScalar}). Here, we also include some additional experimental results for using \texttt{StandardScalar} instead of \texttt{MaxMinScalar}. These are results for \textbf{moderate} changes to the model.

\begin{table}[!htbp]
\caption{Performance on HELOC dataset minimizing for $L_2$ cost.}
\begin{center}
\begin{small}
\begin{sc}
\begin{tabular}{lccc}
\toprule
Method & $L_2$ Cost & Validity & LOF \\
\midrule
CCF    & 3.94& 100.0 \%& 0.96\\
\midrule
FT    & 1.05 & 17.0\%&  0.57 \\
FT +RobX & 2.93  & 94.4\% & 0.95        \\
\midrule
FOCUS    & 1.17 & 30.1\% & 0.69  \\
FOCUS +RobX & 2.94& 97.4\% & 0.96\\
\midrule
FACE  &  5.83 & 90.1\% & 0.74        \\
FACE +RobX &  4.67& 100.0\% & 0.94 \\
\midrule
NN   & 2.71 & 37.6\%& 0.87  \\
NN +RobX  & 3.14 & 98.2\% & 0.98 \\
\bottomrule
\end{tabular}
\end{sc}
\end{small}
\end{center}
\vskip -0.1in
\end{table}

\begin{table}
\caption{Performance on HELOC dataset with $L_2$ cost: FOCUS is applied on the model with a higher threshold, i.e., $M(x) > \gamma$}
\begin{center}
\begin{small}
\begin{sc}
\begin{tabular}{lccc}
\toprule
Method & $L_2$ Cost & Validity & LOF \\
\midrule
FOCUS ($\gamma=$0.5)    & 1.17 & 30.1\% & 0.69  \\
FOCUS ($\gamma=$0.5) +RobX   & 2.94 & 97.4\% & 0.96  \\
\midrule
FOCUS ($\gamma=$0.6) & 1.67& 68.4\% & 0.64\\
FOCUS ($\gamma=$0.6) +RobX & 3.65& 100\% & 0.96\\
\midrule
FOCUS ($\gamma=$0.7) & 2.28& 97.4\% & 0.59\\
FOCUS ($\gamma=$0.7) +RobX & 3.64& 100\% & 0.97\\
\midrule
FOCUS ($\gamma=$0.8) & 3.68& 100\% & 0.55\\
FOCUS ($\gamma=$0.8) +RobX & 3.69& 100\% & 0.55\\
\bottomrule
\end{tabular}
\end{sc}
\end{small}
\end{center}
\vskip -0.1in
\end{table}

\begin{table}
\caption{Performance on German Credit dataset for $L_2$ cost.}
\begin{center}
\begin{small}
\begin{sc}
\begin{tabular}{lccc}
\toprule
Method & $L_2$ Cost & Validity & LOF \\
\midrule
CCF    & 3.18 & 98.2\% & 0.69 \\
\midrule
FT    & 0.56 & 60.0\% &  0.88 \\
FT +RobX & 2.35 & 97.0\% & 0.88       \\
\midrule
FOCUS    & 0.77 & 67.2\%  & 0.87 \\
FOCUS +RobX & 2.83 & 97.0\% & 0.75 \\
\midrule
FACE  & 4.74 & 94.3\% &    0.81    \\
FACE +RobX & 3.38  & 96.9\% & 0.87\\
\midrule
NN   & 1.99 & 76.7 \%& 0.75 \\
NN +RobX  & 2.50 & 96.9\% & 0.81 \\
\bottomrule
\end{tabular}
\end{sc}
\end{small}
\end{center}
\vskip -0.1in
\end{table}

\begin{table}
\caption{Performance on German Credit dataset with $L_2$ cost: FOCUS is applied on the model with a higher threshold, i.e., $M(x) > \gamma$.}
\begin{center}
\begin{small}
\begin{sc}
\begin{tabular}{lcccr}
\toprule
Method & $L_2$ Cost & Validity & LOF \\
\midrule
FOCUS ($\gamma=$0.5)    & 0.77 & 67.2\% & 0.87  \\
FOCUS ($\gamma=$0.5) +RobX   & 2.83 & 97.0\% & 0.75 \\
\midrule
FOCUS ($\gamma=$0.6) & 0.84& 77.6\% & 0.81\\
FOCUS ($\gamma=$0.6) +RobX &2.83 & 97.0 \% & 0.75 \\
\midrule
FOCUS ($\gamma=$0.7) & 0.89 & 82.0\% & 0.81\\
FOCUS ($\gamma=$0.7) +RobX & 2.83& 97.0\% & 0.75\\
\midrule
FOCUS ($\gamma=$0.8) & 1.27& 88.6\% & 0.69\\
FOCUS ($\gamma=$0.8) +RobX & 2.79& 100\% & 0.75\\
\midrule
FOCUS ($\gamma=$0.9) & 1.62& 87.7\% & 0.51\\
FOCUS ($\gamma=$0.9) +RobX & 2.69& 100\% & 0.81\\
\bottomrule
\end{tabular}
\end{sc}
\end{small}
\end{center}
\vskip -0.1in
\end{table}


\end{document}